\documentclass[10pt,twocolumn,letterpaper]{article}

\usepackage{cvpr}
\usepackage{times}
\usepackage{epsfig}
\usepackage{graphicx}
\usepackage{amsmath}
\usepackage{amssymb}

\newcommand{\ve}[1]{\mathbf{#1}} % for displaying a vector
\newcommand{\ma}[1]{\mathrm{#1}} % for displaying a matrix

\newcommand{\var}{\emph{Var}}

\usepackage[british,UKenglish,USenglish,english,american]{babel}

\usepackage{tabulary}

\usepackage{multirow}

\usepackage{hyphenat}

\renewcommand\arraystretch{1.1}

\newcommand{\hide}[1]{}

\usepackage[bookmarks=false,colorlinks=true,linkcolor=black,citecolor=black,filecolor=black,urlcolor=black]{hyperref}

\cvprfinalcopy % *** Uncomment this line for the final submission

\def\cvprPaperID{****} % *** Enter the CVPR Paper ID here

\begin{document}

%%%%%%%%% TITLE
\title{Delving Deep into Rectifiers:\\Surpassing Human-Level Performance on ImageNet Classification}

\author{Kaiming He \qquad Xiangyu Zhang \qquad Shaoqing Ren \qquad Jian Sun \vspace{12pt}\\
Microsoft Research\\
\normalsize
\{kahe,~v-xiangz,~v-shren,~jiansun\}@microsoft.com
}

\maketitle
%\thispagestyle{empty}

%%%%%%%%% ABSTRACT
\begin{abstract}
Rectified activation units (rectifiers) are essential for state-of-the-art neural networks. In this work, we study rectifier neural networks for image classification from two aspects. First, we propose a Parametric Rectified Linear Unit (PReLU) that generalizes the traditional rectified unit. PReLU improves model fitting with nearly zero extra computational cost and little overfitting risk. Second, we derive a robust initialization method that particularly considers the rectifier nonlinearities. This method enables us to train extremely deep rectified models directly from scratch and to investigate deeper or wider network architectures.
Based on our PReLU networks (PReLU-nets), we achieve \textbf{4.94\%} top-5 test error on the ImageNet 2012 classification dataset. This is a 26\% relative improvement over the ILSVRC 2014 winner (GoogLeNet, 6.66\% \cite{Szegedy2014}). To our knowledge, our result is the first to surpass human-level performance (5.1\%, \cite{Russakovsky2014}) on this visual recognition challenge.
\end{abstract}

%%%%%%%%% BODY TEXT
\section{Introduction}

Convolutional neural networks (CNNs) \cite{LeCun1989,Krizhevsky2012} have demonstrated recognition accuracy better than or comparable to humans in several visual recognition tasks, including recognizing traffic signs \cite{Ciresan2012}, faces \cite{Taigman2014,Sun2014}, and hand-written digits \cite{Ciresan2012,Wan2013}. In this work, we present a result that surpasses human-level performance on a more generic and challenging recognition task - the classification task in the 1000-class ImageNet dataset \cite{Russakovsky2014}.

In the last few years, we have witnessed tremendous improvements in recognition performance, mainly due to advances in two technical directions: building more powerful models, and designing effective strategies against overfitting. On one hand, neural networks are becoming more capable of fitting training data, because of increased complexity (\eg, increased depth \cite{Simonyan2014,Szegedy2014}, enlarged width \cite{Zeiler2014,Sermanet2014}, and the use of smaller strides \cite{Zeiler2014,Sermanet2014,Chatfield2014,Simonyan2014}), new nonlinear activations \cite{Nair2010,Maas2013,Zeiler2013,Lin2013,Srivastava2013,Goodfellow2013}, and sophisticated layer designs \cite{Szegedy2014,He2014}. On the other hand, better generalization is achieved by effective regularization techniques \cite{Hinton2012,Srivastava2014,Goodfellow2013,Wan2013}, aggressive data augmentation \cite{Krizhevsky2012,Howard2013,Simonyan2014,Szegedy2014}, and large-scale data \cite{Deng2009,Russakovsky2014}.

Among these advances, the rectifier neuron \cite{Nair2010,Glorot2011,Maas2013,Zeiler2013}, \eg, Rectified Linear Unit (ReLU), is one of several keys to the recent success of deep networks \cite{Krizhevsky2012}. It expedites convergence of the training procedure \cite{Krizhevsky2012} and leads to better solutions \cite{Nair2010,Glorot2011,Maas2013,Zeiler2013} than conventional sigmoid-like units. Despite the prevalence of rectifier networks, recent improvements of models \cite{Zeiler2014,Sermanet2014,He2014,Simonyan2014,Szegedy2014} and theoretical guidelines for training them \cite{Glorot2010,Saxe2013} have rarely focused on the properties of the rectifiers.

In this paper, we investigate neural networks from two aspects particularly driven by the rectifiers.
First, we propose a new generalization of ReLU, which we call \emph{Parametric Rectified Linear Unit} (PReLU). This activation function adaptively learns the parameters of the rectifiers, and improves accuracy at negligible extra computational cost. Second, we study the difficulty of training rectified models that are very deep. By explicitly modeling the nonlinearity of rectifiers (ReLU/PReLU), we derive a theoretically sound initialization method, which helps with convergence of very deep models (\eg, with 30 weight layers) trained directly from scratch. This gives us more flexibility to explore more powerful network architectures.

On the 1000-class ImageNet 2012 dataset, our PReLU network (PReLU-net) leads to a single-model result of 5.71\% top-5 error, which surpasses all existing multi-model results. Further, our multi-model result achieves \textbf{4.94\%} top-5 error on the test set, which is a 26\% relative improvement over the ILSVRC 2014 winner (GoogLeNet, 6.66\% \cite{Szegedy2014}). To the best of our knowledge, our result surpasses for the first time the reported human-level performance (5.1\% in \cite{Russakovsky2014}) on this visual recognition challenge.

\section{Approach}

In this section, we first present the PReLU activation function (Sec.~\ref{sec:prelu}). Then we derive our initialization method for deep rectifier networks (Sec.~\ref{sec:init}). Lastly we discuss our architecture designs (Sec.~\ref{sec:arch}).

\subsection{Parametric Rectifiers}

We show that replacing the parameter-free ReLU activation by a learned parametric activation unit improves classification accuracy\footnote{Concurrent with our work, Agostinelli \etal \cite{Agostinelli2014} also investigated learning activation functions and showed improvement on other tasks.}.

\begin{figure}[t]
\begin{center}
\includegraphics[width=0.85\linewidth]{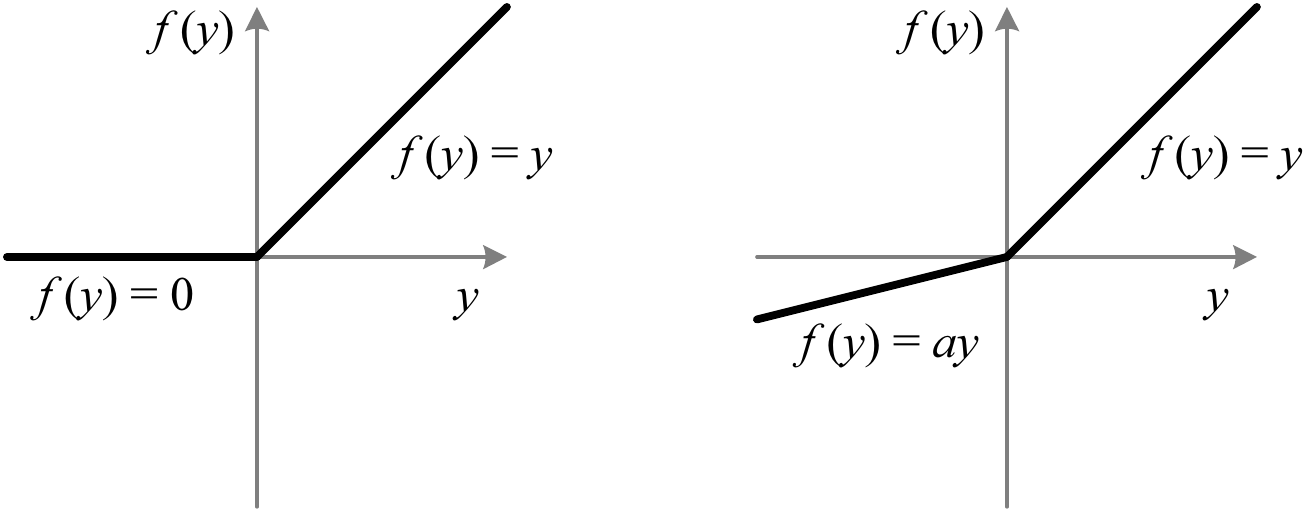}
\end{center}
\caption{ReLU vs. PReLU. For PReLU, the coefficient of the negative part is not constant and is adaptively learned.}
\label{fig:prelu}
\end{figure}

\subsubsection*{Definition}

\label{sec:prelu}
Formally, we consider an activation function defined as:
\begin{equation}\label{eq:prelu}
f(y_i) = \begin{cases} y_i, & \mbox{if } y_i > 0 \\ a_i y_i, & \mbox{if } y_i \leq 0 \end{cases}.
\end{equation}
Here $y_i$ is the input of the nonlinear activation $f$ on the $i$th channel, and $a_i$ is a coefficient controlling the slope of the negative part. The subscript $i$ in $a_i$ indicates that we allow the nonlinear activation to vary on different channels. When $a_i = 0$, it becomes ReLU; when $a_i$ is a learnable parameter, we refer to Eqn.(\ref{eq:prelu}) as \emph{Parametric ReLU} (PReLU). Figure~\ref{fig:prelu} shows the shapes of ReLU and PReLU.
Eqn.(\ref{eq:prelu}) is equivalent to $f(y_i) = \max(0, y_i)+a_i\min(0, y_i)$.

If $a_i$ is a small and fixed value, PReLU becomes the Leaky ReLU (LReLU) in \cite{Maas2013} ($a_i=0.01$). The motivation of LReLU is to avoid zero gradients. Experiments in \cite{Maas2013} show that LReLU has negligible impact on accuracy compared with ReLU. On the contrary, our method adaptively learns the PReLU parameters jointly with the whole model. We hope for end-to-end training that will lead to more specialized activations.

PReLU introduces a very small number of extra parameters. The number of extra parameters is equal to the total number of channels, which is negligible when considering the total number of weights. So we expect no extra risk of overfitting. We also consider a channel-shared variant: $f(y_i) = \max(0, y_i) + a \min(0, y_i)$ where the coefficient is shared by all channels of one layer. This variant only introduces a single extra parameter into each layer.

\subsubsection*{Optimization}

PReLU can be trained using backpropagation \cite{LeCun1989} and optimized simultaneously with other layers.
The update formulations of $\{a_i\}$ are simply derived from the chain rule. The gradient of $a_i$ for one layer is:
\begin{equation}\label{eq:grad}
\frac{\partial \mathcal{E}}{\partial a_i} = \sum_{y_i} \frac{\partial \mathcal{E}}{\partial f(y_i)}\frac{\partial f(y_i)}{\partial a_i},
\end{equation}
where $\mathcal{E}$ represents the objective function. The term $\frac{\partial \mathcal{E}}{\partial f(y_i)}$ is the gradient propagated from the deeper layer.
The gradient of the activation is given by:
\begin{equation}
\frac{\partial f(y_i)}{\partial a_i} = \begin{cases} 0, & \mbox{if } y_i > 0 \\ y_i, & \mbox{if } y_i \leq 0 \end{cases}.
\end{equation}
The summation $\sum_{y_i}$ runs over all positions of the feature map. For the channel-shared variant, the gradient of $a$ is $\frac{\partial \mathcal{E}}{\partial a} = \sum_i\sum_{y_i} \frac{\partial \mathcal{E}}{\partial f(y_i)}\frac{\partial f(y_i)}{\partial a}$, where $\sum_i$ sums over all channels of the layer. The time complexity due to PReLU is negligible for both forward and backward propagation.

We adopt the momentum method when updating $a_i$:
\begin{equation}\label{eq:update}
\Delta a_i := \mu \Delta a_i + \epsilon \frac{\partial \mathcal{E}}{\partial a_i}.
\end{equation}
Here $\mu$ is the momentum and $\epsilon$ is the learning rate. It is worth noticing that we do not use weight decay ($l_2$ regularization) when updating $a_i$. A weight decay tends to push $a_i$ to zero, and thus biases PReLU toward ReLU. Even without regularization, the learned coefficients rarely have a magnitude larger than 1 in our experiments.
Further, we do not constrain the range of $a_i$ so that the activation function may be non-monotonic. We use $a_i=0.25$ as the initialization throughout this paper.

\subsubsection*{Comparison Experiments}

We conducted comparisons on a deep but efficient model with 14 weight layers. The model was studied in \cite{He2014a} (model E of \cite{He2014a}) and its architecture is described in Table~\ref{tab:s14}. We choose this model because it is sufficient for representing a category of very deep models, as well as to make the experiments feasible.

As a baseline, we train this model with ReLU applied in the convolutional (conv) layers and the first two fully-connected (fc) layers. The training implementation follows \cite{He2014a}. The top-1 and top-5 errors are 33.82\% and 13.34\% on ImageNet 2012, using 10-view testing (Table~\ref{tab:s14_ablated}).

\setlength{\tabcolsep}{4pt}
\begin{table}[t]
\begin{center}
\footnotesize
\begin{tabular}{c c |c|c}
\hline
\multicolumn{2}{c|}{ } & \multicolumn{2}{c}{learned coefficients} \\
\hline
\multicolumn{2}{c|}{layer} &\footnotesize{channel-shared} & \footnotesize{channel-wise} \\
\hline
\hline
conv1 &  7$\times$7, 64, $_{/2}$ & 0.681 & 0.596 \\
\hline
pool1 &  3$\times$3, $_{/3}$ & & \\
\hline
conv2$_{1}$ &  2$\times$2, 128 & 0.103 & 0.321 \\
conv2$_{2}$ &  2$\times$2, 128 & 0.099 & 0.204 \\
conv2$_{3}$ &  2$\times$2, 128 & 0.228 & 0.294 \\
conv2$_{4}$ &  2$\times$2, 128 & 0.561 & 0.464 \\
\hline
pool2 &  2$\times$2, $_{/2}$ & & \\
\hline
conv3$_{1}$ &  2$\times$2, 256 & 0.126 & 0.196 \\
conv3$_{2}$ &  2$\times$2, 256 & 0.089 & 0.152 \\
conv3$_{3}$ &  2$\times$2, 256 & 0.124 & 0.145 \\
conv3$_{4}$ &  2$\times$2, 256 & 0.062 & 0.124 \\
conv3$_{5}$ &  2$\times$2, 256 & 0.008 & 0.134 \\
conv3$_{6}$ &  2$\times$2, 256 & 0.210 & 0.198 \\
\hline
spp &  $\{6,3,2,1\}$ & & \\
\hline
fc$_{1}$ &  4096 & 0.063 & 0.074 \\
fc$_{2}$ &  4096 & 0.031 & 0.075 \\
fc$_{3}$ &  1000 &       &       \\
\hline
\end{tabular}
\end{center}
\caption{A small but deep 14-layer model \cite{He2014a}. The filter size and filter number of each layer is listed. The number {/$s$} indicates the stride $s$ that is used.
The learned coefficients of PReLU are also shown. For the channel-wise case, the average of $\{a_i\}$ over the channels is shown for each layer.}
\label{tab:s14}
\end{table}

Then we train the same architecture from scratch, with all ReLUs replaced by PReLUs (Table~\ref{tab:s14_ablated}). The top-1 error is reduced to 32.64\%. This is a \textbf{1.2\%} gain over the ReLU baseline. Table~\ref{tab:s14_ablated} also shows that channel-wise/channel-shared PReLUs perform comparably. For the channel-shared version, PReLU only introduces 13 extra free parameters compared with the ReLU counterpart. But this small number of free parameters play critical roles as evidenced by the 1.1\% gain over the baseline. This implies the importance of adaptively learning the shapes of activation functions.

\setlength{\tabcolsep}{6pt}
\begin{table}[t]
\begin{center}
\small
\begin{tabular}{c|cc}
\hline
   & top-1 & top-5 \\
\hline
\hline
ReLU & 33.82 & 13.34\\
\hline
PReLU, channel-shared & 32.71 & 12.87\\
PReLU, channel-wise   & \textbf{32.64} & \textbf{12.75}\\
\hline
\end{tabular}
\end{center}
\caption{Comparisons between ReLU and PReLU on the small model. The error rates are for ImageNet 2012 using 10-view testing. The images are resized so that the shorter side is 256, during both training and testing. Each view is 224$\times$224. All models are trained using 75 epochs.}
\label{tab:s14_ablated}
\end{table}

Table~\ref{tab:s14} also shows the learned coefficients of PReLUs for each layer.
There are two interesting phenomena in Table~\ref{tab:s14}. First, the first conv layer (conv1) has coefficients (0.681 and 0.596) significantly greater than 0. As the filters of conv1 are mostly Gabor-like filters such as edge or texture detectors, the learned results show that both positive and negative responses of the filters are respected. We believe that this is a more economical way of exploiting low-level information, given the limited number of filters (\eg, 64). Second, for the channel-wise version, the deeper conv layers in general have smaller coefficients. This implies that the activations gradually become ``more nonlinear'' at increasing depths. In other words, the learned model tends to keep more information in earlier stages and becomes more discriminative in deeper stages.

\subsection{Initialization of Filter Weights for Rectifiers}
\label{sec:init}

Rectifier networks are easier to train \cite{Glorot2011,Krizhevsky2012,Zeiler2013} compared with traditional sigmoid-like activation networks. But a bad initialization can still hamper the learning of a highly non-linear system. In this subsection, we propose a robust initialization method that removes an obstacle of training extremely deep rectifier networks.

Recent deep CNNs are mostly initialized by random weights drawn from Gaussian distributions \cite{Krizhevsky2012}.
With fixed standard deviations (\eg, 0.01 in \cite{Krizhevsky2012}),
very deep models (\eg, $>$8 conv layers) have difficulties to converge, as reported by the VGG team \cite{Simonyan2014} and also observed in our experiments. To address this issue, in \cite{Simonyan2014} they pre-train a model with 8 conv layers to initialize deeper models. But this strategy requires more training time, and may also lead to a poorer local optimum. In \cite{Szegedy2014,Lee2014}, auxiliary classifiers are added to intermediate layers to help with convergence.

Glorot and Bengio \cite{Glorot2010} proposed to adopt a properly scaled uniform distribution for initialization. This is called ``\emph{Xavier}'' initialization in \cite{Jia2014}. Its derivation is based on the assumption that the activations are linear. This assumption is invalid for ReLU and PReLU.

In the following, we derive a theoretically more sound initialization by taking ReLU/PReLU into account. In our experiments, our initialization method allows for extremely deep models (\eg, 30 conv/fc layers) to converge, while the ``\emph{Xavier}'' method \cite{Glorot2010} cannot.

\subsubsection*{Forward Propagation Case}

Our derivation mainly follows \cite{Glorot2010}. The central idea is to investigate the variance of the responses in each layer.

For a conv layer, a response is:
\begin{equation}\label{eq:forward}
\ve{y}_l=\ma{W}_l\ve{x}_l + \ve{b}_l.
\end{equation}
Here, $\ve{x}$ is a $k^2c$-by-1 vector that
represents co-located $k$$\times$$k$ pixels in $c$ input channels. $k$ is the spatial filter size of the layer. With $n=k^2c$ denoting the number of connections of a response, $\ma{W}$ is a $d$-by-$n$ matrix, where $d$ is the number of filters and each row of $\ma{W}$ represents the weights of a filter. $\ve{b}$ is a vector of biases, and $\ve{y}$ is the response at a pixel of the output map.
We use $l$ to index a layer. We have
$\ve{x}_{l}=f(\ve{y}_{l-1})$
where $f$ is the activation. We also have $c_l = d_{l-1}$. 

We let the initialized elements in $\ma{W}_{l}$ be mutually independent and share the same distribution.
As in \cite{Glorot2010}, we assume that the elements in $\ve{x}_l$ are also mutually independent and share the same distribution, and $\ve{x}_l$ and $\ma{W}_{l}$ are independent of each other. Then we have:
\begin{equation}
\var[y_{l}]=n_l\var[w_{l}x_l],
\end{equation}
where now $y_{l}$, $x_{l}$, and $w_{l}$ represent the random variables of each element in $\ve{y}_l$, $\ma{W}_l$, and $\ve{x}_l$ respectively. We let $w_{l}$ have zero mean. Then the variance of the product of independent variables gives us:
\begin{equation}\label{eq:y1}
\var[y_{l}]=n_l\var[w_{l}]E[x^2_{l}].
\end{equation}
Here $E[x^2_{l}]$ is the expectation of the square of $x_l$. It is worth noticing that $E[x^2_{l}]\neq \var[x_l]$ unless $x_l$ has zero mean. For the ReLU activation, $x_{l}=max(0, y_{l-1})$ and thus it does not have zero mean. This will lead to a conclusion different from \cite{Glorot2010}.

If we let $w_{l-1}$ have a symmetric distribution around zero and $b_{l-1}=0$, then $y_{l-1}$ has zero mean and has a symmetric distribution around zero. This leads to
$E[x^2_{l}]=\frac{1}{2}\var[y_{l-1}]$ when $f$ is ReLU.
Putting this into Eqn.(\ref{eq:y1}), we obtain:
\begin{equation}\label{eq:y2}
\var[y_{l}]=\frac{1}{2}n_l\var[w_{l}]\var[y_{l-1}].
\end{equation}
With $L$ layers put together, we have:
\begin{equation}\label{eq:prod_fw}
\var[y_{L}]=\var[y_{1}]\left(\prod_{l=2}^{L}\frac{1}{2}n_l\var[w_{l}]\right).
\end{equation}
This product is the key to the initialization design. A proper initialization method should avoid reducing or magnifying the magnitudes of input signals exponentially. So we expect the above product to take a proper scalar (\eg, 1). A sufficient condition is:
\begin{equation}\label{eq:init_fw}
\frac{1}{2}n_l\var[w_{l}]=1, \quad \forall l.
\end{equation}
This leads to a zero-mean Gaussian distribution whose standard deviation (std) is $\sqrt{2/{n_l}}$. This is our way of initialization. We also initialize $\ve{b}=0$.

For the first layer ($l=1$), we should have $n_1\var[w_{1}]=1$ because there is no ReLU applied on the input signal.
But the factor $1/2$ does not matter if it just exists on one layer. So we also adopt Eqn.(\ref{eq:init_fw}) in the first layer for simplicity.

\subsubsection*{Backward Propagation Case}

For back-propagation, the gradient of a conv layer is computed by:
\begin{equation}\label{eq:backward}
\Delta \ve{x}_l = \ma{\hat{W}}_l \Delta \ve{y}_l.
\end{equation}
Here we use $\Delta \ve{x}$ and $\Delta \ve{y}$ to denote gradients ($\frac{\partial\mathcal{E}}{\partial\ve{x}}$ and $\frac{\partial\mathcal{E}}{\partial\ve{y}}$) for simplicity.
$\Delta \ve{y}$ represents $k$-by-$k$ pixels in $d$ channels, and is reshaped into a $k^2d$-by-1 vector. We denote $\hat{n}=k^2d$. Note that $\hat{n}\neq n=k^2c$. $\ma{\hat{W}}$ is a $c$-by-$\hat{n}$ matrix where the filters are rearranged in the way of back-propagation. Note that $\ma{W}$ and $\ma{\hat{W}}$ can be reshaped from each other. $\Delta \ve{x}$ is a $c$-by-1 vector representing the gradient at a pixel of this layer. As above, we assume that $w_l$ and $\Delta y_l$ are independent of each other, then $\Delta x_l$ has zero mean for all $l$, when $w_l$ is initialized by a symmetric distribution around zero.

In back-propagation we also have $\Delta y_{l} = f'(y_l)\Delta x_{l+1}$ where $f'$ is the derivative of $f$. For the ReLU case, $f'(y_l)$ is zero or one, and their probabilities are equal. We assume that $f'(y_l)$ and $\Delta x_{l+1}$ are independent of each other. Thus we have $E[\Delta y_l]=E[\Delta x_{l+1}]/2=0$, and also $E[(\Delta y_l)^2]=\var[\Delta y_l]=\frac{1}{2}\var[\Delta x_{l+1}]$.
Then we compute the variance of the gradient in Eqn.(\ref{eq:backward}):
\begin{eqnarray}
\var[\Delta x_l] &=& \hat{n}_l\var[w_l]\var[\Delta y_l]\nonumber\\
 &=&\frac{1}{2}\hat{n}_l\var[w_l]\var[\Delta x_{l+1}].\label{eq:dx2}
\end{eqnarray}
The scalar $1/2$ in both Eqn.(\ref{eq:dx2}) and Eqn.(\ref{eq:y2}) is the result of ReLU, though the derivations are different.
With $L$ layers put together, we have:
\begin{equation}\label{eq:prod_bw}
\var[\Delta x_2] = \var[\Delta x_{L+1}]\left(\prod_{l=2}^{L}\frac{1}{2}\hat{n}_{l}\var[w_{l}]\right).
\end{equation}
We consider a sufficient condition that the gradient is not exponentially large/small:
\begin{equation}\label{eq:init_bw}
\frac{1}{2}\hat{n}_l\var[w_{l}]=1, \quad \forall l.
\end{equation}
The only difference between this equation and Eqn.(\ref{eq:init_fw}) is that $\hat{n}_l=k_l^2d_l$ while $n_l=k_l^2c_l=k_l^2d_{l-1}$. Eqn.(\ref{eq:init_bw}) results in a zero-mean Gaussian distribution whose std is $\sqrt{2/{\hat{n}_l}}$.

For the first layer ($l=1$), we need not compute $\Delta x_1$ because it represents the image domain. But we can still adopt Eqn.(\ref{eq:init_bw}) in the first layer, for the same reason as in the forward propagation case - the factor of a single layer does not make the overall product exponentially large/small.

We note that it is sufficient to use either Eqn.(\ref{eq:init_bw}) or Eqn.(\ref{eq:init_fw}) alone. For example, if we use Eqn.(\ref{eq:init_bw}), then in Eqn.(\ref{eq:prod_bw}) the product $\prod_{l=2}^{L}\frac{1}{2}\hat{n}_{l}\var[w_{l}]=1$, and in Eqn.(\ref{eq:prod_fw}) the product $\prod_{l=2}^{L}\frac{1}{2}n_l\var[w_{l}]=\prod_{l=2}^{L}n_l/\hat{n}_{l}=c_{2}/d_{L}$, which is not a diminishing number in common network designs. This means that if the initialization properly scales the backward signal, then this is also the case for the forward signal; and vice versa.
For all models in this paper, both forms can make them converge.

\begin{figure}[t]
\begin{center}
\includegraphics[width=0.8\linewidth]{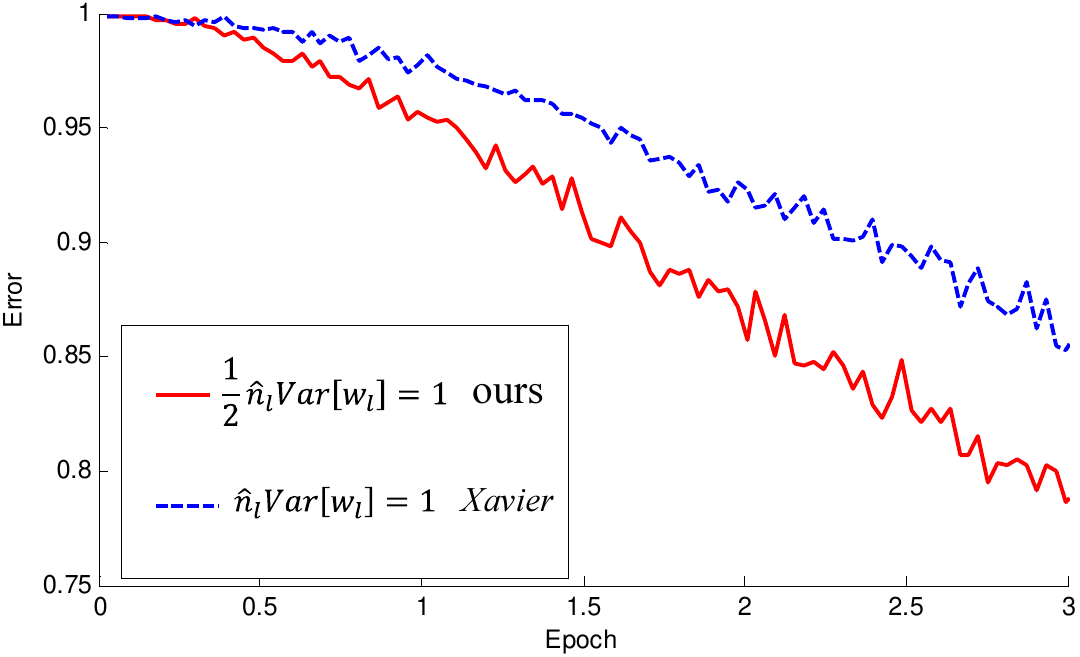}
\end{center}
\caption{The convergence of a \textbf{22-layer} large model (B in Table~\ref{tab:arch}). The x-axis is the number of training epochs. The y-axis is the top-1 error of 3,000 random val samples, evaluated on the center crop. We use ReLU as the activation for both cases. Both our initialization (red) and ``\emph{Xavier}'' (blue) \cite{Glorot2010} lead to convergence, but ours starts reducing error earlier.}
\label{fig:converge_22layers}
%\end{figure}
%\begin{figure}[t]
\begin{center}
\includegraphics[width=0.8\linewidth]{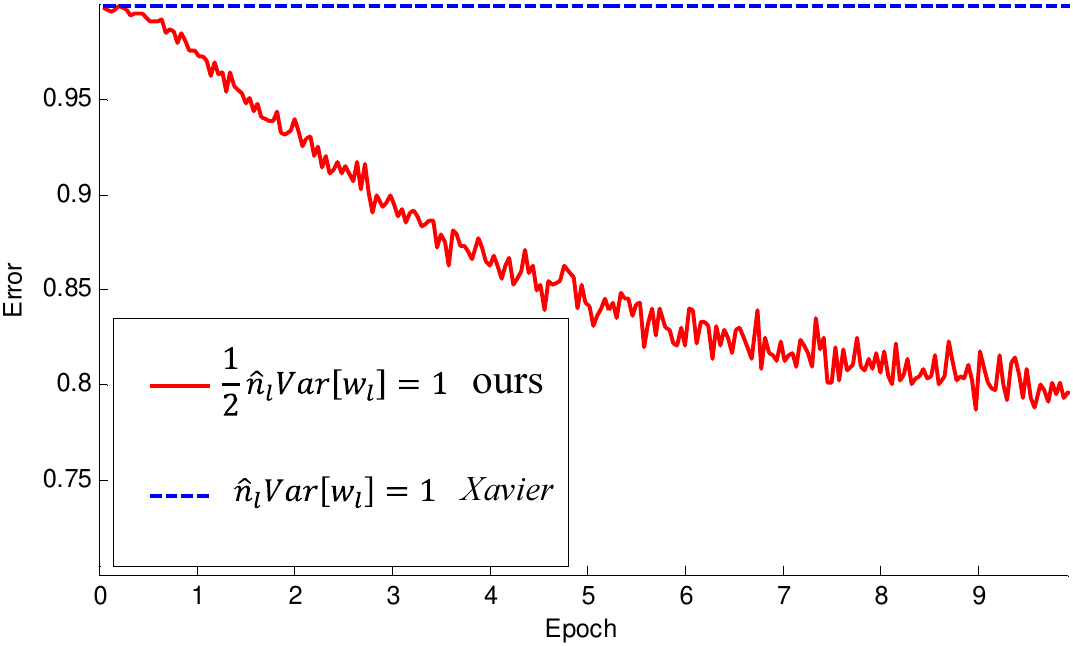}
\end{center}
\caption{The convergence of a \textbf{30-layer} small model (see the main text). We use ReLU as the activation for both cases. Our initialization (red) is able to make it converge. But ``\emph{Xavier}'' (blue) \cite{Glorot2010} completely stalls - we also verify that its gradients are all diminishing. It does not converge even given more epochs.}
\label{fig:converge_30layers}
\end{figure}

\subsubsection*{Discussions}

If the forward/backward signal is inappropriately scaled by a factor $\beta$ in each layer, then the final propagated signal will be rescaled by a factor of $\beta^L$ after $L$ layers, where $L$ can represent some or all layers. When $L$ is large, if $\beta>1$, this leads to extremely amplified signals and an algorithm output of infinity; if $\beta<1$, this leads to diminishing signals\footnote{In the presence of weight decay (l$_2$ regularization of weights), when the gradient contributed by the logistic loss function is diminishing, the total gradient is not diminishing because of the weight decay. A way of diagnosing diminishing gradients is to check whether the gradient is modulated only by weight decay.}. In either case, the algorithm does not converge - it diverges in the former case, and stalls in the latter.

Our derivation also explains why the constant standard deviation of 0.01 makes some deeper networks stall \cite{Simonyan2014}. We take ``model B'' in the VGG team's paper \cite{Simonyan2014} as an example. This model has 10 conv layers all with 3$\times$3 filters. The filter numbers ($d_l$) are 64 for the 1st and 2nd layers, 128 for the 3rd and 4th layers, 256 for the 5th and 6th layers, and 512 for the rest.
The std computed by Eqn.(\ref{eq:init_bw}) ($\sqrt{2/{\hat{n}_l}}$) is 0.059, 0.042, 0.029, and 0.021 when the filter numbers are 64, 128, 256, and 512 respectively. If the std is initialized as 0.01, the std of the gradient propagated from conv10 to conv2 is $1/(5.9\times4.2^2\times2.9^2\times2.1^4)=1/(1.7\times10^4)$ of what we derive. This number may explain why diminishing gradients were observed in experiments.

It is also worth noticing that the variance of the input signal can be roughly preserved from the first layer to the last. In cases when the input signal is not normalized (\eg, it is in the range of $[-128, 128]$), its magnitude can be so large that the softmax operator will overflow. A solution is to normalize the input signal, but this may impact other hyper-parameters. Another solution is to include a small factor on the weights among all or some layers, \eg, $\sqrt[L]{1/128}$ on $L$ layers. In practice, we use a std of 0.01 for the first two fc layers and 0.001 for the last. These numbers are smaller than they should be (\eg, $\sqrt{2/4096}$) and will address the normalization issue of images whose range is about $[-128, 128]$.

For the initialization in the PReLU case, it is easy to show that Eqn.(\ref{eq:init_fw}) becomes:
\begin{equation}\label{eq:init_fw_prelu}
\frac{1}{2}(1+a^2)n_l\var[w_{l}]=1, \quad \forall l,
\end{equation}
where $a$ is the initialized value of the coefficients. If $a=0$, it becomes the ReLU case; if $a=1$, it becomes the linear case (the same as \cite{Glorot2010}). Similarly, Eqn.(\ref{eq:init_bw}) becomes $\frac{1}{2}(1+a^2)\hat{n}_l\var[w_{l}]=1$.

\subsubsection*{Comparisons with ``\emph{Xavier}'' Initialization \cite{Glorot2010}}

The main difference between our derivation and the ``\emph{Xavier}'' initialization \cite{Glorot2010} is that we address the rectifier nonlinearities\footnote{There are other minor differences. In \cite{Glorot2010}, the derived variance is adopted for uniform distributions, and the forward and backward cases are averaged. But it is straightforward to adopt their conclusion for Gaussian distributions and for the forward or backward case only.}. The derivation in \cite{Glorot2010} only considers the linear case, and its result is given by $n_l\var[w_{l}]=1$ (the forward case), which can be implemented as a zero-mean Gaussian distribution whose std is $\sqrt{1/{n_l}}$. When there are $L$ layers, the std will be ${1}/{\sqrt{2}^L}$ of our derived std. This number, however, is not small enough to completely stall the convergence of the models actually used in our paper (Table~\ref{tab:arch}, up to 22 layers) as shown by experiments.
Figure~\ref{fig:converge_22layers} compares the convergence of a 22-layer model.
Both methods are able to make them converge. But ours starts reducing error earlier. We also investigate the possible impact on accuracy. For the model in Table~\ref{tab:s14_ablated} (using ReLU), the ``\emph{Xavier}'' initialization method leads to 33.90/13.44 top-1/top-5 error, and ours leads to 33.82/13.34. We have not observed clear superiority of one to the other on accuracy.

Next, we compare the two methods on extremely deep models with up to 30 layers (27 conv and 3 fc). We add up to sixteen conv layers with 256 2$\times$2 filters in the model in Table~\ref{tab:s14}. Figure~\ref{fig:converge_30layers} shows the convergence of the 30-layer model. Our initialization is able to make the extremely deep model converge. On the contrary, the ``\emph{Xavier}'' method completely stalls the learning, and the gradients are diminishing as monitored in the experiments.

These studies demonstrate that we are ready to investigate extremely deep, rectified models by using a more principled initialization method. But in our current experiments on ImageNet, we have not observed the benefit from training extremely deep models. For example, the aforementioned 30-layer model has 38.56/16.59 top-1/top-5 error, which is clearly worse than the error of the 14-layer model in Table~\ref{tab:s14_ablated} (33.82/13.34).
Accuracy saturation or degradation was also observed in the study of small models \cite{He2014a},
VGG's large models \cite{Simonyan2014}, and in speech recognition \cite{Zeiler2013}. This is perhaps because the method of increasing depth is not appropriate, or the recognition task is not enough complex.

Though our attempts of extremely deep models have not shown benefits, 
our initialization method paves a foundation for further study on increasing depth. We hope this will be helpful in other more complex tasks.

\renewcommand\arraystretch{1.04}
\setlength{\tabcolsep}{8pt}
\begin{table*}[t]
\begin{center}
%\footnotesize
\small
\begin{tabular}{|c||c||c|c|c|}
\hline
input size & VGG-19 \cite{Simonyan2014} & model A & model B  & model C \\
\hline
\multirow{3}{*}{224} &  3$\times$3, 64 & 7$\times$7, 96, /2 & 7$\times$7, 96, /2 & 7$\times$7, 96, /2\\
                     &  3$\times$3, 64 &                    &                    &                   \\
                     &  2$\times$2 maxpool, /2 &                 &                    &                   \\
\hline
\multirow{3}{*}{112} &  3$\times$3, 128 &                   &                    &                   \\
                     &  3$\times$3, 128 &                   &                    &                   \\
                     &  2$\times$2 maxpool, /2 & 2$\times$2 maxpool, /2 & 2$\times$2 maxpool, /2  & 2$\times$2 maxpool, /2\\
\hline
\multirow{7}{*}{56} &  3$\times$3, 256 &  3$\times$3, 256 &  3$\times$3, 256  &  3$\times$3, 384\\
                    &  3$\times$3, 256 &  3$\times$3, 256 &  3$\times$3, 256  &  3$\times$3, 384\\
                    &  3$\times$3, 256 &  3$\times$3, 256 &  3$\times$3, 256  &  3$\times$3, 384\\
                    &  3$\times$3, 256 &  3$\times$3, 256 &  3$\times$3, 256  &  3$\times$3, 384\\
                    &                  &  3$\times$3, 256 &  3$\times$3, 256  &  3$\times$3, 384\\
                    &                  &                  &  3$\times$3, 256  &  3$\times$3, 384\\
                    &  2$\times$2 maxpool, /2 & 2$\times$2 maxpool, /2 & 2$\times$2 maxpool, /2 & 2$\times$2 maxpool, /2\\
\hline
\multirow{7}{*}{28} &  3$\times$3, 512 &  3$\times$3, 512 &  3$\times$3, 512  &  3$\times$3, 768\\
                    &  3$\times$3, 512 &  3$\times$3, 512 &  3$\times$3, 512  &  3$\times$3, 768\\
                    &  3$\times$3, 512 &  3$\times$3, 512 &  3$\times$3, 512  &  3$\times$3, 768\\
                    &  3$\times$3, 512 &  3$\times$3, 512 &  3$\times$3, 512  &  3$\times$3, 768\\
                    &                  &  3$\times$3, 512 &  3$\times$3, 512  &  3$\times$3, 768\\
                    &                  &                  &  3$\times$3, 512  &  3$\times$3, 768\\
                    &  2$\times$2 maxpool, /2 & 2$\times$2 maxpool, /2 & 2$\times$2 maxpool, /2 & 2$\times$2 maxpool, /2\\
\hline
\multirow{7}{*}{14} &  3$\times$3, 512 &  3$\times$3, 512 &  3$\times$3, 512  &  3$\times$3, 896\\
                    &  3$\times$3, 512 &  3$\times$3, 512 &  3$\times$3, 512  &  3$\times$3, 896\\
                    &  3$\times$3, 512 &  3$\times$3, 512 &  3$\times$3, 512  &  3$\times$3, 896\\
                    &  3$\times$3, 512 &  3$\times$3, 512 &  3$\times$3, 512  &  3$\times$3, 896\\
                    &                  &  3$\times$3, 512 &  3$\times$3, 512  &  3$\times$3, 896\\
                    &                  &                  &  3$\times$3, 512  &  3$\times$3, 896\\
                    &  2$\times$2 maxpool, /2   &  spp, $\{7,3,2,1\}$ & spp, $\{7,3,2,1\}$ & spp, $\{7,3,2,1\}$ \\
\hline
fc$_{1}$ &  \multicolumn{4}{c|}{4096} \\
fc$_{2}$ &  \multicolumn{4}{c|}{4096} \\
fc$_{3}$ &  \multicolumn{4}{c|}{1000} \\
\hline
depth (conv+fc) & 19 & 19 & 22 & 22\\
\hline
complexity (ops., $\times 10^{10}$)     & 1.96 & 1.90 & 2.32 & 5.30\\
\hline
\end{tabular}
\end{center}
\caption{Architectures of large models. Here ``/2'' denotes a stride of 2.}
\label{tab:arch}
\end{table*}
\renewcommand\arraystretch{1.1}

\subsection{Architectures}
\label{sec:arch}

The above investigations provide guidelines of designing our architectures, introduced as follows.

Our baseline is the 19-layer model (A) in Table~\ref{tab:arch}. For a better comparison, we also list the VGG-19 model \cite{Simonyan2014}. Our model A has the following modifications on VGG-19: (i) in the first layer, we use a filter size of 7$\times$7 and a stride of 2; (ii) we move the other three conv layers on the two largest feature maps (224, 112) to the smaller feature maps (56, 28, 14). The time complexity (Table~\ref{tab:arch}, last row) is roughly unchanged because the deeper layers have more filters; (iii) we use spatial pyramid pooling (SPP) \cite{He2014} before the first fc layer. The pyramid has 4 levels - the numbers of bins are 7$\times$7, 3$\times$3, 2$\times$2, and 1$\times$1, for a total of 63 bins.

It is worth noticing that we have no evidence that our model A is a better \emph{architecture} than VGG-19, though our model A has better results than VGG-19's result reported by \cite{Simonyan2014}. In our earlier experiments with less scale augmentation, we observed that our model A and our reproduced VGG-19 (with SPP and our initialization) are comparable. The main purpose of using model A is for faster running speed. The actual running time of the conv layers on larger feature maps is slower than those on smaller feature maps, when their time complexity is the same. In our four-GPU implementation, our model A takes 2.6s per mini-batch (128), and our reproduced VGG-19 takes 3.0s, evaluated on four Nvidia K20 GPUs.

In Table~\ref{tab:arch}, our model B is a deeper version of A. It has three extra conv layers. Our model C is a wider (with more filters) version of B. The width substantially increases the complexity, and its time complexity is about 2.3$\times$ of B (Table~\ref{tab:arch}, last row). Training A/B on four K20 GPUs, or training C on eight K40 GPUs, takes about 3-4 weeks.

We choose to increase the model width instead of depth, because deeper models have only diminishing improvement or even degradation on accuracy.
In recent experiments on small models \cite{He2014a}, it has been found that aggressively increasing the depth leads to saturated or degraded accuracy. In the VGG paper \cite{Simonyan2014}, the 16-layer and 19-layer models perform comparably. In the speech recognition research of \cite{Zeiler2013}, the deep models degrade when using more than 8 hidden layers (all being fc).
We conjecture that similar degradation may also happen on larger models for ImageNet. We have monitored the training procedures of some extremely deep models (with 3 to 9 layers added on B in Table~\ref{tab:arch}), and found both training and testing error rates degraded in the first 20 epochs (but we did not run to the end due to limited time budget, so there is not yet solid evidence that these large and overly deep models will ultimately degrade). Because of the possible degradation, we choose not to further increase the depth of these large models.

On the other hand, the recent research \cite{Eigen2013} on small datasets suggests that the accuracy should improve from the increased number of parameters in conv layers. This number depends on the depth and width. So we choose to increase the width of the conv layers to obtain a higher-capacity model.

While all models in Table~\ref{tab:arch} are very large, we have not observed severe overfitting. We attribute this to the aggressive data augmentation used throughout the whole training procedure, as introduced below.

\section{Implementation Details}

\subsubsection*{Training}

Our training algorithm mostly follows \cite{Krizhevsky2012,Howard2013,Chatfield2014,He2014,Simonyan2014}. From a resized image whose shorter side is $s$, a 224$\times$224 crop is randomly sampled, with the per-pixel mean subtracted. The scale $s$ is randomly jittered in the range of $[256, 512]$, following \cite{Simonyan2014}. One half of the random samples are flipped horizontally \cite{Krizhevsky2012}. Random color altering \cite{Krizhevsky2012} is also used.

Unlike \cite{Simonyan2014} that applies scale jittering only during fine-tuning, we apply it from the beginning of training. Further, unlike \cite{Simonyan2014} that initializes a deeper model using a shallower one, we directly train the very deep model using our initialization described in Sec.~\ref{sec:init} (we use Eqn.(\ref{eq:init_bw})). Our end-to-end training may help improve accuracy, because it may avoid poorer local optima.

Other hyper-parameters that might be important are as follows. The weight decay is 0.0005, and momentum is 0.9. Dropout (50\%) is used in the first two fc layers. The mini-batch size is fixed as 128.
The learning rate is 1e-2, 1e-3, and 1e-4, and is switched when the error plateaus.
The total number of epochs is about 80 for each model.

\subsubsection*{Testing}

We adopt the strategy of ``multi-view testing on feature maps'' used in the SPP-net paper \cite{He2014}. We further improve this strategy using the dense sliding window method in \cite{Sermanet2014,Simonyan2014}.

We first apply the convolutional layers on the resized full image and obtain the last convolutional feature map. In the feature map, each 14$\times$14 window is pooled using the SPP layer \cite{He2014}. The fc layers are then applied on the pooled features to compute the scores. This is also done on the horizontally flipped images. The scores of all dense sliding windows are averaged \cite{Sermanet2014,Simonyan2014}.
We further combine the results at multiple scales as in \cite{He2014}.

\subsubsection*{Multi-GPU Implementation}

We adopt a simple variant of Krizhevsky's method \cite{Krizhevsky2014} for parallel training on multiple GPUs.
We adopt ``data parallelism'' \cite{Krizhevsky2014} on the conv layers. The GPUs are synchronized before the first fc layer. Then the forward/backward propagations of the fc layers are performed on a single GPU - this means that we do not parallelize the computation of the fc layers. The time cost of the fc layers is low, so it is not necessary to parallelize them. This leads to a simpler implementation than the ``model parallelism'' in \cite{Krizhevsky2014}.
Besides, model parallelism introduces some overhead due to the communication of filter responses, and is not faster than computing the fc layers on just a single GPU.

We implement the above algorithm on our modification of the Caffe library \cite{Jia2014}. 
We do not increase the mini-batch size (128) because the accuracy may be decreased \cite{Krizhevsky2014}. For the large models in this paper, we have observed a 3.8x speedup using 4 GPUs, and a 6.0x speedup using 8 GPUs.

\section{Experiments on ImageNet}

We perform the experiments on the 1000-class ImageNet 2012 dataset \cite{Russakovsky2014} which contains about 1.2 million training images, 50,000 validation images, and 100,000 test images (with no published labels). The results are measured by top-1/top-5 error rates \cite{Russakovsky2014}. We only use the provided data for training. All results are evaluated on the validation set, except for the final results in Table~\ref{tab:ensemble}, which are evaluated on the test set. The top-5 error rate is the metric officially used to rank the methods in the classification challenge \cite{Russakovsky2014}.

\setlength{\tabcolsep}{12pt}
\begin{table}[t]
\begin{center}
\small
\begin{tabular}{c|cc|cc}
\hline
 model A & \multicolumn{2}{c|}{ReLU} & \multicolumn{2}{c}{PReLU} \\
\hline
  scale $s$  & top-1 & top-5 & top-1 & top-5 \\
\hline
\hline
256 & 26.25 & 8.25 & \textbf{25.81} & \textbf{8.08} \\
384 & 24.77 & 7.26 & \textbf{24.20} & \textbf{7.03} \\
480 & 25.46 & 7.63 & \textbf{24.83} & \textbf{7.39} \\
\hline
multi-scale & 24.02 & 6.51 & \textbf{22.97} & \textbf{6.28} \\
\hline
\end{tabular}
\end{center}
\caption{Comparisons between ReLU/PReLU on model A in ImageNet 2012 using dense testing.}
\label{tab:model_a}
\end{table}

\subsubsection*{Comparisons between ReLU and PReLU}

In Table~\ref{tab:model_a}, we compare ReLU and PReLU on the large model A. We use the channel-wise version of PReLU. For fair comparisons, both ReLU/PReLU models are trained using the same total number of epochs, and the learning rates are also switched after running the same number of epochs.

Table~\ref{tab:model_a} shows the results at three scales and the multi-scale combination. The best single scale is 384, possibly because it is in the middle of the jittering range $[256, 512]$.
For the multi-scale combination, PReLU reduces the top-1 error by 1.05\% and the top-5 error by 0.23\% compared with ReLU.
The results in Table~\ref{tab:s14_ablated} and Table~\ref{tab:model_a} consistently show that PReLU improves both small and large models. This improvement is obtained with almost no computational cost.

\subsubsection*{Comparisons of Single-model Results}

Next we compare single-model results. We first show 10-view testing results \cite{Krizhevsky2012} in Table~\ref{tab:10view}. Here, each view is a 224-crop. The 10-view results of VGG-16 are based on our testing using the publicly released model \cite{Simonyan2014} as it is not reported in \cite{Simonyan2014}. Our best 10-view result is 7.38\% (Table~\ref{tab:10view}). Our other models also outperform the existing results.

Table~\ref{tab:single} shows the comparisons of single-model results, which are all obtained using multi-scale and multi-view (or dense) test. Our results are denoted as MSRA. Our baseline model (A+ReLU, 6.51\%) is already substantially better than the best existing single-model result of 7.1\% reported for VGG-19 in the latest update of \cite{Simonyan2014} (arXiv v5). We believe that this gain is mainly due to our end-to-end training, without the need of pre-training shallow models.

Moreover, our best single model (C, PReLU) has \textbf{5.71\%} top-5 error. This result is even better than all previous multi-model results (Table~\ref{tab:ensemble}).
Comparing A+PReLU with B+PReLU, we see that the 19-layer model and the 22-layer model perform comparably. On the other hand, increasing the width (C \vs B, Table~\ref{tab:single}) can still improve accuracy. This indicates that when the models are deep enough, the width becomes an essential factor for accuracy.

\renewcommand\arraystretch{1.15}
\setlength{\tabcolsep}{6pt}
\begin{table*}[t]
\begin{center}
\small
\begin{tabular}{c|cc}
\hline
 model & top-1 & top-5 \\
\hline
\hline
MSRA \cite{He2014} & 29.68 & 10.95 \\
VGG-16 \cite{Simonyan2014} & 28.07$^{\dag}$ & 9.33$^{\dag}$\\
GoogLeNet \cite{Szegedy2014} & - & 9.15 \\
\hline
A, ReLU    & 26.48 & 8.59 \\
A, PReLU  & 25.59 & 8.23 \\
B, PReLU  & 25.53 & 8.13 \\
C, PReLU  & \textbf{24.27} & \textbf{7.38}  \\
\hline
\end{tabular}
\end{center}
\caption{The single-model \textbf{10-view} results for ImageNet 2012 val set.~$^{\dag}$: Based on our tests.}
\label{tab:10view}
%\end{table*}
%\vspace{8pt}
\newcommand{\tabincell}[2]{\begin{tabular}{@{}#1@{}}#2\end{tabular}}
\setlength{\tabcolsep}{16pt}
%\begin{table*}[t]
\small
\begin{center}
\begin{tabular}{c|c|c c}
\hline
  & team & top-1 & top-5\\
\hline
\hline
\multirow{3}{*}{\tabincell{c}{ in competition \\ ILSVRC 14}} &  MSRA \cite{He2014} & 27.86 & 9.08$^{\dag}$\\
                              &  VGG \cite{Simonyan2014} & - & 8.43$^{\dag}$\\
                              &  GoogLeNet \cite{Szegedy2014} & - & 7.89\\
\hline
\multirow{7}{*}{post-competition} & VGG \cite{Simonyan2014} (arXiv v2) & 24.8~~ & 7.5~~\\
                              &  VGG \cite{Simonyan2014} (arXiv v5) & 24.4~~ & 7.1~~\\
                              &  Baidu \cite{Wu2015} & 24.88 & 7.42\\\cline{2-4}
                              &  MSRA (A, ReLU) & 24.02 & 6.51 \\
                              &  MSRA (A, PReLU) & 22.97 & 6.28 \\
                              &  MSRA (B, PReLU) & 22.85 & 6.27 \\
                              &  MSRA (C, PReLU) & \textbf{21.59} & \textbf{5.71} \\
\hline
\end{tabular}
\end{center}
\caption{The \textbf{single-model} results for ImageNet 2012 val set.~$^{\dag}$: Evaluated from the test set.}
\label{tab:single}
%\end{table*}
%\vspace{8pt}
\setlength{\tabcolsep}{16pt}
%\begin{table*}[t]
\small
\begin{center}
\begin{tabular}{c|c|c}
\hline
  & team & top-5 (\textbf{test}) \\
\hline
\hline
\multirow{3}{*}{\tabincell{c}{ in competition \\ ILSVRC 14}} &  MSRA, SPP-nets \cite{He2014} & 8.06 \\
                              &  VGG \cite{Simonyan2014} & 7.32\\
                              &  GoogLeNet \cite{Szegedy2014} & 6.66\\
\hline
\multirow{3}{*}{post-competition} & VGG \cite{Simonyan2014} (arXiv v5) & 6.8 \\
                              &  Baidu \cite{Wu2015} & 5.98 \\
                              &  \textbf{MSRA, PReLU-nets} & \textbf{4.94} \\
\hline
\end{tabular}
\end{center}
\caption{The \textbf{multi-model} results for the ImageNet 2012 test set.}
\label{tab:ensemble}
\end{table*}

\subsubsection*{Comparisons of Multi-model Results}

We combine six models including those in Table~\ref{tab:single}. For the time being we have trained only one model with architecture C. The other models have accuracy inferior to C by considerable margins. We conjecture that we can obtain better results by using fewer stronger models.

The multi-model results are in Table~\ref{tab:ensemble}. Our result is \textbf{4.94\%} top-5 error on the test set.
This number is evaluated by the ILSVRC server, because the labels of the test set are not published.
Our result is 1.7\% better than the ILSVRC 2014 winner (GoogLeNet, 6.66\% \cite{Szegedy2014}), which represents a $\sim$26\% relative improvement. This is also a $\sim$17\% relative improvement over the latest result (Baidu, 5.98\% \cite{Wu2015}).

\begin{figure}[t]
\begin{center}
\includegraphics[width=1.0\linewidth]{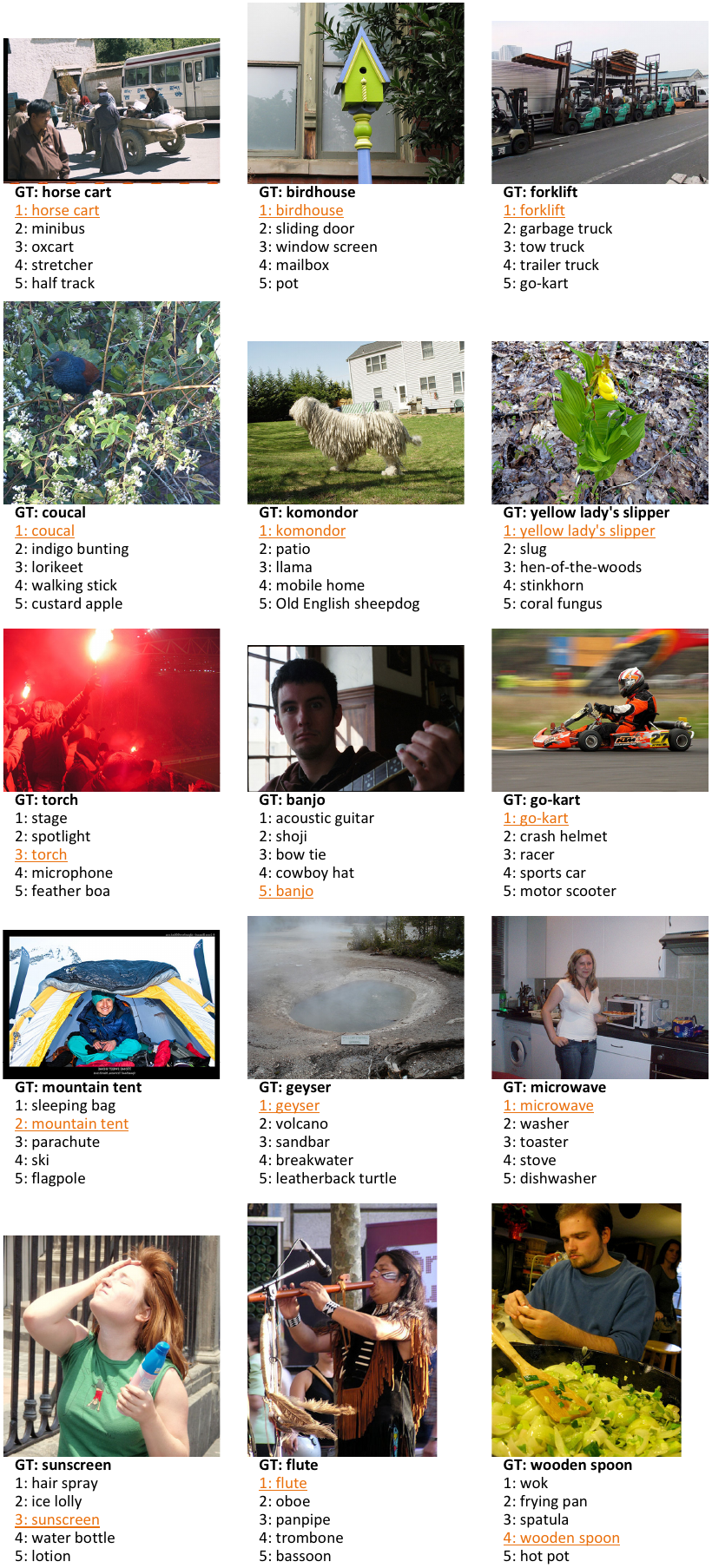}
\end{center}
\caption{Example validation images successfully classified by our method. For each image, the ground-truth label and the top-5 labels predicted by our method are listed.}
\label{fig:good_cases}
\end{figure}

\subsubsection*{Analysis of Results}

Figure~\ref{fig:good_cases} shows some example validation images successfully classified by our method. Besides the correctly predicted labels, we also pay attention to the other four predictions in the top-5 results. Some of these four labels are other objects in the multi-object images, \eg, the ``horse-cart'' image (Figure~\ref{fig:good_cases}, row 1, col 1) contains a ``mini-bus'' and it is also recognized by the algorithm. Some of these four labels are due to the uncertainty among similar classes, \eg, the ``coucal'' image (Figure~\ref{fig:good_cases}, row 2, col 1) has predicted labels of other bird species.

\begin{figure}[t]
\begin{center}
\includegraphics[width=1.0\linewidth]{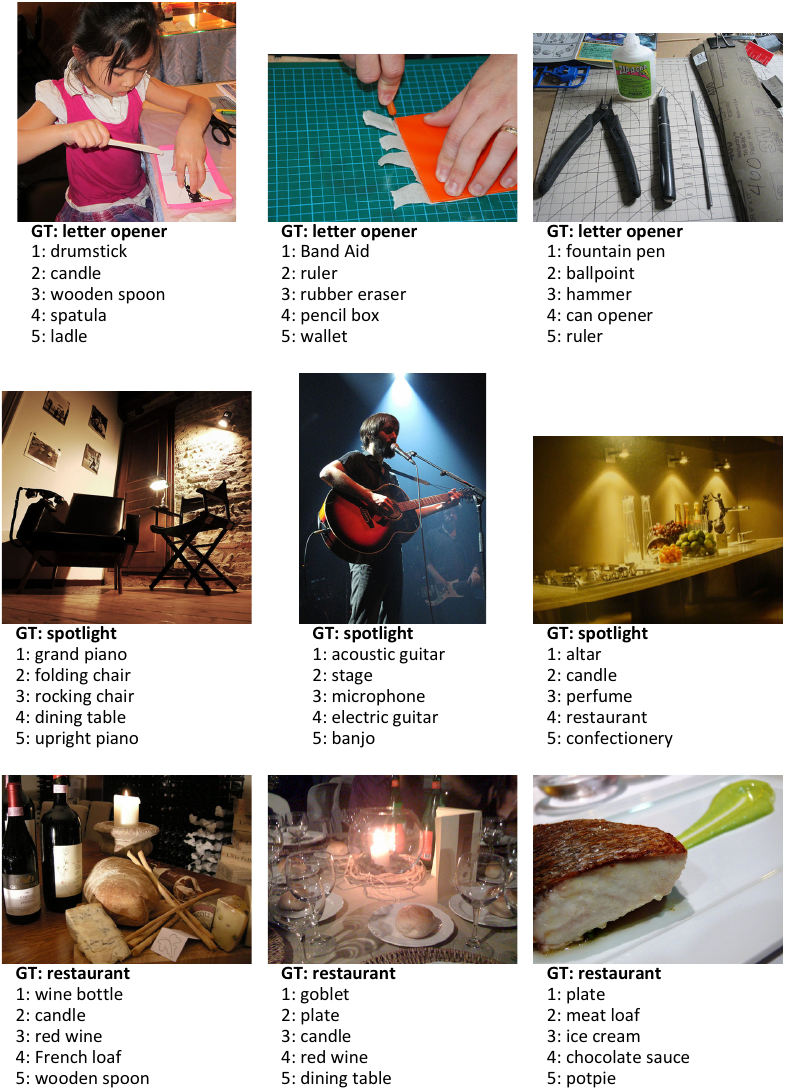}
\end{center}
\caption{Example validation images incorrectly classified by our method, in the three classes with the highest top-5 test error. Top: ``letter opener'' (49\% top-5 test error). Middle: ``spotlight'' (38\%). Bottom: ``restaurant'' (36\%). For each image, the ground-truth label and the top-5 labels predicted by our method are listed.}
\label{fig:bad_cases}
\end{figure}

\begin{figure}[t]
\begin{center}
\includegraphics[width=.96\linewidth]{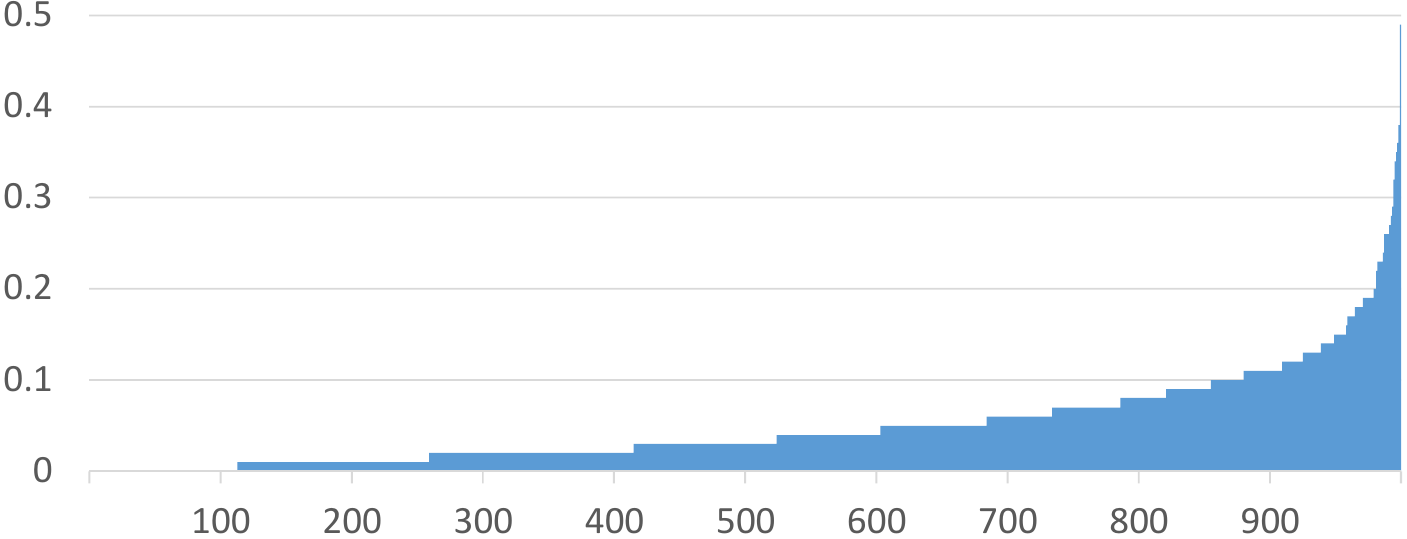}
\end{center}
\caption{The per-class top-5 errors of our result (average of 4.94\%) on the test set. Errors are displayed in ascending order.}
\label{fig:error}
%\end{figure}
%\begin{figure}[t]
\begin{center}
\includegraphics[width=.96\linewidth]{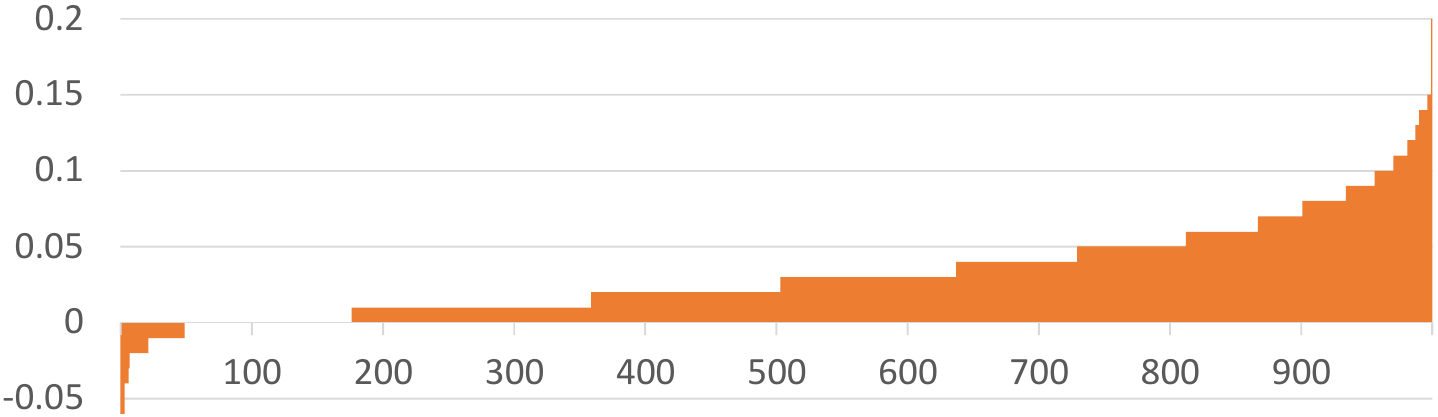}
\end{center}
\caption{The difference of top-5 error rates between our result (average of 4.94\%) and our team's in-competition result for ILSVRC 2014 (average of 8.06\%) on the test set, displayed in ascending order. A positive number indicates a reduced error rate.}
\label{fig:error_reduce}
\end{figure}

Figure~\ref{fig:error} shows the per-class top-5 error of our result (average of 4.94\%) on the test set, displayed in ascending order. Our result has zero top-5 error in 113 classes - the images in these classes are all correctly classified. The three classes with the highest top-5 error are ``letter opener'' (49\%), ``spotlight'' (38\%), and ``restaurant'' (36\%). The error is due to the existence of multiple objects, small objects, or large intra-class variance. Figure~\ref{fig:bad_cases} shows some example images misclassified by our method in these three classes. Some of the predicted labels still make some sense.

In Figure~\ref{fig:error_reduce}, we show the per-class difference of top-5 error rates between our result (average of 4.94\%) and our team's in-competition result in ILSVRC 2014 (average of 8.06\%). The error rates are reduced in 824 classes, unchanged in 127 classes, and increased in 49 classes.

\subsubsection*{Comparisons with Human Performance from \cite{Russakovsky2014}}

Russakovsky \etal \cite{Russakovsky2014} recently reported that human performance yields a 5.1\% top-5 error on the ImageNet dataset. This number is achieved by a human annotator who is well trained on the validation images to be better aware of the existence of relevant classes. When annotating the test images, the human annotator is given a special interface, where each class title is accompanied by a row of 13 example training images. The reported human performance is estimated on a random subset of 1500 test images.

Our result (4.94\%) exceeds the reported human-level performance. To our knowledge, our result is the first published instance of surpassing humans on this visual recognition challenge. The analysis in \cite{Russakovsky2014} reveals that the two major types of human errors come from fine-grained recognition and class unawareness. The investigation in \cite{Russakovsky2014} suggests that algorithms can do a better job on fine-grained recognition (\eg, 120 species of dogs in the dataset). The second row of Figure~\ref{fig:good_cases} shows some example fine-grained objects successfully recognized by our method - ``coucal'', ``komondor'', and ``yellow lady's slipper''. While humans can easily recognize these objects as a bird, a dog, and a flower, it is nontrivial for most humans to tell their species.
On the negative side, our algorithm still makes mistakes in cases that are not difficult for humans, especially for those requiring context understanding or high-level knowledge (\eg, the ``spotlight'' images in Figure~\ref{fig:bad_cases}).

While our algorithm produces a superior result on this particular dataset, this does not indicate that machine vision outperforms human vision on object recognition in general.
On recognizing elementary object categories (\ie, common objects or concepts in daily lives) such as the Pascal VOC task \cite{Everingham2010}, machines still have obvious errors in cases that are trivial for humans.
Nevertheless, we believe that our results show the tremendous potential of machine algorithms to match human-level performance on visual recognition.

{\small
\bibliographystyle{ieee}
\bibliography{prelu}
}

\end{document}